\documentclass[10pt,conference]{IEEEtran}
\usepackage[utf8]{inputenc}
\usepackage{graphicx}
\usepackage{hyperref} 
\usepackage{color}
\usepackage{authblk}
\usepackage{algorithmicx}
\usepackage[ruled,noend,linesnumbered]{algorithm2e}
\SetKwRepeat{Do}{do}{while}
\usepackage[noend]{algpseudocode}
\usepackage{multirow}
\usepackage{graphicx}
\usepackage{subfigure}

\usepackage{listings}
\usepackage{fancyhdr}
\usepackage{amssymb}

\newcommand{\commentout}[1]{}

\newcommand{\rnnfunction}{\psi}

\newcommand{\real}{\mathbb{R}}

\newcommand{\testRNN}{\textbf{testRNN}}

\newcommand{\param}[1]{\texttt{#1}}
\newcommand{\num}[2]{$#1$$,$$#2$}

\title{\testRNN: Coverage-guided Testing on Recurrent Neural Networks}
\author[1]{Wei Huang}
\author[2]{Youcheng Sun}
\author[3]{James Sharp}
\author[1]{Xiaowei Huang}
\affil[1]{University of Liverpool, UK}
\affil[2]{Queen's University Belfast, UK}
\affil[3]{Defence Science and Technology Laboratory (Dstl), UK}

\begin{document}

\maketitle
\begin{abstract}
Recurrent neural networks (RNNs) have been widely applied to various sequential tasks such as text processing, video recognition, and molecular property prediction. We introduce the \emph{first} coverage-guided testing tool, \textit{coined} \testRNN,  for the verification and validation of a major class of RNNs, long short-term memory networks (LSTMs). The tool implements a generic mutation-based test case generation method, and it empirically evaluates the robustness of a network using three novel LSTM structural test coverage metrics.
Moreover, it is able to help the model designer go through the internal data flow processing of the LSTM layer. 
The tool is available through: \url{https://github.com/TrustAI/testRNN} under the  BSD 3-Clause licence.
\end{abstract}

\section{introduction}

Development of recurrent neural networks (RNNs) has been focused predominantly on improving empirical accuracy,
and far less has been done towards their verification and validation. Verification and validation (V\&V) are independent procedures that are used together for checking that a product, service, or system meets requirements and specifications and that it fulfills its intended purpose \cite{GHTF2004}. 
There is a clear need therefore
to move towards developing techniques to validate RNNs against their specifications. We focus on a major class of RNNs, i.e., long short-term memory networks (LSTMs), and an important specification, 
that is, their
robustness. For robustness, it is required that a prediction made by an LSTM is invariant with respect to small perturbations of the input \cite{szegedy2014intriguing}.

Our approach is based on coverage-guided testing \cite{YLW2006}, which has been shown successful in software fault detection. Coverage-guided testing has been extended to work with feed-forward neural networks (FNNs) in recent work such as \cite{PCYJ2017,wicker2018feature}, where a collection of test metrics and test case generation algorithms are proposed. These metrics are based on the structural information of the FNNs, such as the neurons \cite{PCYJ2017,ma2018deepgauge}, the relation between neurons in  neighboring layers \cite{sun2018testing,sun2018concolic}, etc. However, when working with RNNs, whose internal structures are much more intricate, new test metrics and new test case generation methods are needed to take into account the additional structures and complexity.

In \cite{HSHS2019}, three LSTM structural test metrics and a mutation-based test case generation method are proposed.
In this work, we develop \testRNN, a testing and debugging tool specially designed for RNNs,
implementing the testing method from \cite{HSHS2019}. 
The test metrics are designed to exploit different functional components of the LSTM networks, by considering both the one-step information change and the multi-step information evolution; the information is extracted from gate vectors and hidden state vectors. To 
refrain from ``gaming against criteria'', the test case generation avoids using 
test metrics as targets, as recommended by Chilensky and Miller in~
\cite{MCDC}.  In \cite{DXLMZL2019}, RNNs are abstracted into state machines, based on which quantitative analysis metrics are applied. The metrics in \cite{DXLMZL2019} are not based on the structural information of RNNs, and the testing tool is not released.

\testRNN\ is useful for both model designers and model users. Before putting their LSTM models into practical use, designers can apply coverage-guided testing to find adversarial examples, which can be used for model improvement via e.g., data augmentation. For users, they may refer to \testRNN\ as a useful tool to see if the LSTM applications are sufficiently robust for the usage in critical circumstances. 

\section{preliminaries}

\subsection{Recurrent Neural Networks (RNNs)}


RNNs work with sequential input data, and consists of at least one recurrent layer. The recurrent layer
can be represented as a function $\rnnfunction:X'\times C\times Y' \rightarrow C\times Y'$ such that $\rnnfunction(x_t,c_{t-1},h_{t-1})=(c_t,h_t)$ for $t=1...n$, where $t$ denotes the $t$-th time step, $c_t$ is the cell state and acts as the intermediate memory, $x_t$ is the input, and $h_t$ is the output.   Intuitively, the recurrent layer takes as inputs the current time input $x_t$, the previous time memory state $c_{t-1}$ and the previous time output   $h_{t-1}$, updates the memory state into $c_t$, and returns $h_{t}$ as the current time output.   Initially, we let $c_0$ and $h_0$ be 0-valued vectors. For a (finite) sequence of inputs $x_1,...,x_n$, the function $\rnnfunction$ is applied recursively over these inputs. 
For example, the popular long short-term memory (LSTM) layer can be represented with the following equations for time~$t$: 
\begin{equation}
\begin{array}{lcl}
f_t & = & \sigma(W_f\cdot [h_{t-1},x_t] + b_f) \\ 
i_t & = & \sigma(W_i\cdot [h_{t-1},x_t] + b_i) \\ 
c_t & = & f_t*c_{t-1} + i_t * tanh(W_c\cdot [h_{t-1},x_t] + b_c)\\
o_t & = & \sigma(W_o\cdot [h_{t-1},x_t] + b_o) \\ 
h_t & = &  o_t * tanh(c_t)
\end{array}
\label{equ:lstm}
\end{equation}
where $ \sigma $ is the sigmoid function such that $\sigma (x) \in [0,1] $ for any $x \in \real $, $ tanh $ is the hyperbolic tangent function such that $\tanh (x) \in  [-1,1] $ for any $x \in \real $, $ W_f, W_i, W_c, W_o $, which are weight matrices, $ b_f, b_i, b_c, b_o $ are 
bias vectors, $ f_t, i_t, o_t $ are internal gate variables, $h_t$ is the hidden state variable (utilising $ o_t $), and $ c_t $ is the cell state variable. 
Normally, a fully-connected layer will receive the outputs $h_t$ from an RNN layer and further process these for the final classifications or predictions.

\subsection{Internal Information of LSTM Cells}

Coverage metrics in \cite{HSHS2019} are based on the structural information extracted from the basic working units, i.e., cells of LSTM. Equations in (\ref{equ:lstm}) represent a set of mathematics operations inside a cell at time step $t$. Based on them, the following information is captured in each cell. 
\paragraph{Aggregate information of hidden states}

The component value of 
$h_t$ varies between -1 and 1.
It is thus reasonable to divide the possible information of the $i^{th}$ component $h_t(i)$ into positive ($>0$) and negative ($<0$). For 
the vector 
$h_t$, we take the aggregate information \cite{tang2017memory} $\xi_t = (\xi_t^+, \xi_t^-)$ by 
summing up 
positive  and negative component values, respectively, such that 
\begin{equation}
\begin{array}{lcl}
\xi_t^+  =  \displaystyle \sum \{ h_t(i) ~|~ i \in \{1..|h_t|\}, h_t(i) > 0 \}  \\

\xi_t^-  =  \displaystyle \sum \{ h_t(i) ~|~ i \in \{1..|h_t|\}, h_t(i) < 0 \}
\end{array}
\end{equation}
Intuitively, $\xi_t^+$ represents the extent to which the hidden state $h_t$ contains positive information and $\xi_t^-$ represents the extent to which the hidden state  $h_t$ contains negative information. To compare the hidden states of adjacent cells, we have
\begin{equation}\label{equ:deltaggregate}
\Delta\xi_t=|\xi_{t}^+-\xi_{t-1}^+|+|\xi_{t}^--\xi_{t-1}^-|
\end{equation} 
which compares the information of the current step $\xi_t$ with its previous step $\xi_{t-1}$.

\paragraph{Abstract information of gates}
We also consider information represented in the gates $f$, $i$, and $o$. Intuitively, these gates have their dedicated meanings. That is, the forget gate, $f$, controls to what extent information passes through the cell, the input gate, $i$, determines how much new information will be added to the cell state, and the output gate, $o$, determines 
the level of impact the cell state has on the hidden output. 

Let $f_{x,t}$ be the value of the gate $f$ at time $t$ when the input is $x$.  We use forget rate $R_t(f,x)$ to denote the extent to which information passes through the cell, at time $t$, to be memorised, when the input is $x$. 
Formally, we have 
\begin{equation}
R_t(f,x)=
\frac{1}{|f_{x,t}|}\sum_{i=1}^{|f_{x,t}|} f_{x,t}(i).
\end{equation}
It is easy to see that $R_t(f,x)$ is within the range $[0,1]$, since all components in $f_{x,t}$ have their values bounded in $[0,1]$.

\section{The \testRNN\ Tool}
\begin{figure}
    \centering
    \includegraphics[width=0.8\linewidth]{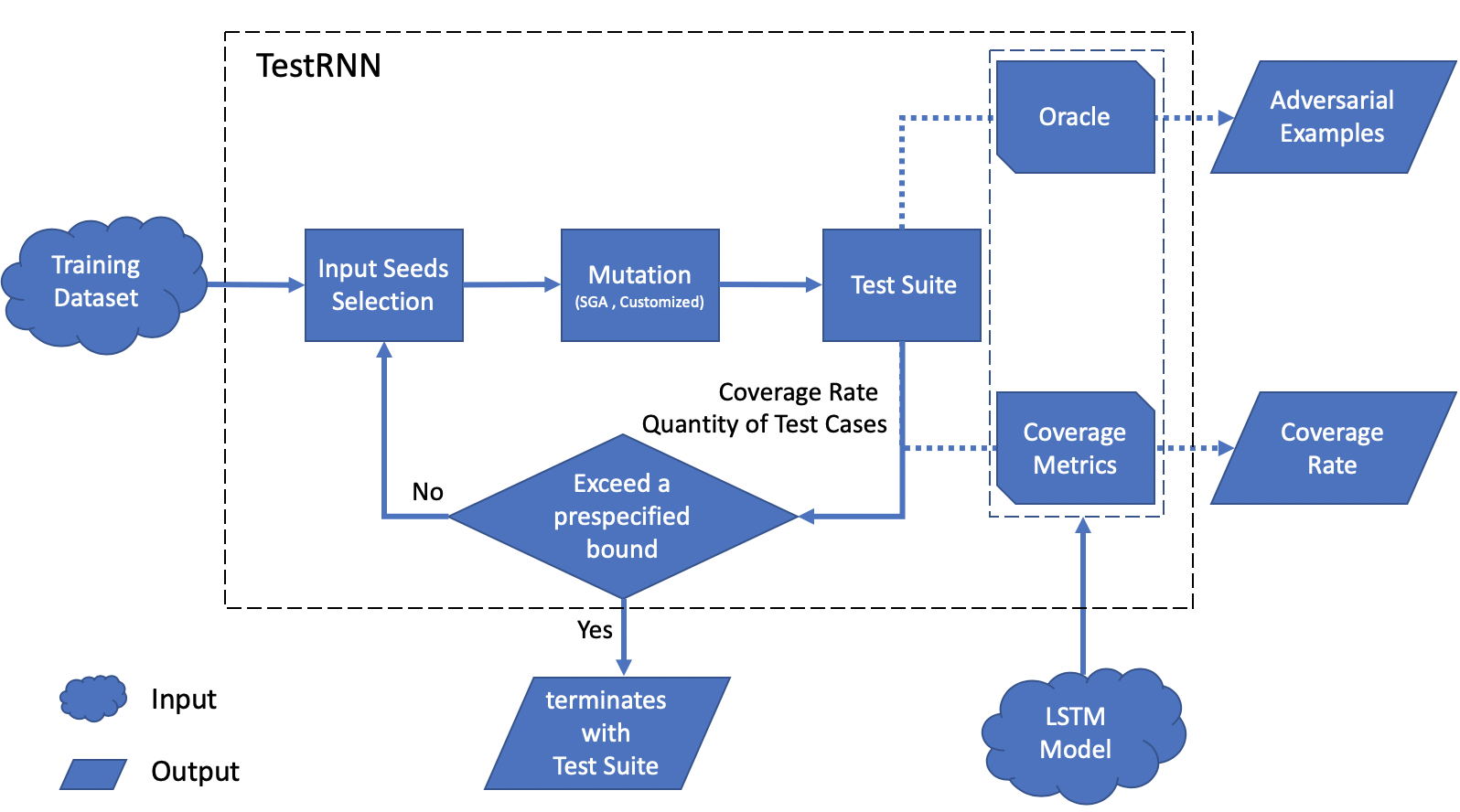}
    \caption{Architecture of the RNN testing tool: \testRNN}
    \label{fig:arch}
    \vspace{-10pt}
\end{figure}
The architecture of \testRNN\ is shown in Figure \ref{fig:arch}. Given an input \param{LSTM Model}, and the 
\param{Training Dataset}, \testRNN\ will generate a \param{Test Suite}, together with a test report that logs the update of 
\param{Coverage Rate}, \param{Adversarial Examples}, and detailed testing information including the 
average perturbation of adversarial examples. 


\subsection{Test Metrics}

\testRNN\ currently supports three structure-based test metrics \cite{HSHS2019} to exploit the behaviours of a LSTM model: \textbf{cell coverage}, \textbf{gate coverage} and \textbf{sequence coverage}. 

Cell coverage aims at covering significant hidden state changes $\Delta\xi_t$ at each time step.
When a cell value $\Delta\xi_t$ is greater than 
$\alpha_h$, a user defined threshold parameter, the cell is activated and covered by the test case. The coverage is then used
to measure the percentage of cells activated at least once by the generated test cases.

The gate coverage is similar to cell coverage, but instead information is extracted from the gates of a LSTM cell. In particular, our tool focuses on the forget gate coverage. The forget gate value $R_t(f,x)$ indicates how much information can be inherited from the last cell. Since LSTM is well known for its long term memory ability, it is meaningful to check if a cell throws away a suitable amount of  information learned from previous inputs. When using forget gate coverage, a threshold value $\alpha_f$ is thus required.

Sequence coverage captures the sequential information passing through the hidden states. It 
comprises positive sequence coverage, based on $\xi_t^+$, and negative sequence coverage, based on $\xi_t^-$. 
A sequence such as $\xi_1^+\xi_2^+...\xi_n^+$ is contained in $n$ symbols. 
In \testRNN, the user is able to decide the number of symbols to be used in the symbolic representation. 
We use both 2 and 3 in our experiments. Based on symbolic representation, we can collect a set of 
layer-wise memory patterns from a set of test cases, and define the coverage rate as the percentage of possible patterns 
covered by the generated test cases. 
Since the number of patterns increases exponentially with respect to the length of sequence, we allow the users to define a specific range of cells to be considered. For example, one could take the range $[451,500]$ within the full range $[1,500]$ to study the last 50 time steps of LSTM processing. 

\subsection{Mutation-based Test Case Generation}
The \param{Mutation} module in Figure \ref{fig:arch} implements the test generation algorithm in \cite{HSHS2019}.
It mutates the input dataset for a higher coverage rate and production of more adversarial examples. There are two kinds of mutations: 
continuous input mutation, and discrete input mutation.

For continuous inputs, such as images, the tool provides a Stochastic Gradient Ascent (SGA) engine. The engine is in principle gradient search \cite{goodfellow2014explaining}, mutating
an input based on two parameters: a gradient magnitude $\epsilon$, and a gradient steps $\tau$, both of which are randomly selected from pre-specified ranges. Thus, it is easy to use SGA to generate multiple mutations from one input seed from the training dataset.

For discrete input problems, a series of  customised mutation operations is very often needed. \testRNN\ also provides several default mutation
methods for commonly-seen discrete  problems. For example, for Natural Language Processing (NLP), available mutation operations  includes (1) synonym replacement, (2) random insertion, (3) random swap, (4) random deletion. 

\subsection{Parameter Settings}
In addition to setting thresholds for test metrics, the user should specify either a level of coverage rate, or a number of test cases, as the stopping criterion. These are required input parameters for the tool.
The user can change the pseudo-random number in the \param{Input Seeds Selection} module to select different sets of seeds from the training database. The minimal test suite generation is optional for \param{Test Suite}. If it is not specified, all the mutation samples will be added into the final test suite.


\subsection{Oracle}

Similar as in \cite{wicker2018feature,sun2018testing,ma2018deepgauge}, an oracle is employed to automatically determine if a mutated test case
is meaningful. We consider an oracle as a set of Euclidean norm balls with the seed inputs as centers. %
Only those test cases that fall within such norm balls are valid, among which the misclassified ones are considered 
faulty, i.e., adversarial examples \cite{szegedy2014intriguing}. 
The radius of norm balls is configurable.

\subsection{Generated Test Suite}

The tool \testRNN\ outputs a test suite that also includes adversarial examples. 
In addition, the user can ask for a minimal test suite to satisfy the specified test coverage. 

\subsection{Coverage Report and Empirical Robustness Evaluation}
 
A coverage report tracks the update of all coverage statistics during the test case generation. The quantity and quality of adversarial examples are also recorded. The quality of adversarial examples is measured by the average perturbation to the original inputs. \testRNN\ includes a post-processing script (namely \param{readfile.py}) that reads the report and automatically extracts the running statistics into a \param{.MAT} file, to be plotted via Matlab. 

The coverage rate and the percentage of adversarial examples in the test suite generated are 
empirical statistics for the robustness evaluation, under the specified coverage metric. Finally, users can (1) compare different coverage metrics for the same model and (2) analyse the impact of threshold values on the coverage rates with the help of plotted figures.

\subsection{Additional usage}
\testRNN\ provides a lot of useful features for LSTM testing. As an example, it records the number of times a test-condition is covered, according to the generated test suite. In Section~\ref{sec:experiments}, we show that a plot on coverage times can help 
users understand the learning mechanism behind the LSTM model.

In conclusion, there are at least the following advantages for using \testRNN. 
First, it generates test cases and can find a number of adversarial examples complying with the specified test-conditions. Second, users  can use \testRNN\ to compare the robustness of different LSTMs. Third, the adversarial examples can be utilized to retrain or improve a model. Last but not least, the plots on coverage times can give intuitive explanations on how a model learns from a dataset.

\section{Usage Example}

\testRNN\ is written in Python. It provides users a 
range of convenient command line options. 
An example is given below.
\begin{verbatim}
python main.py --model network
--TestCaseNum 2000 --threshold_CC 6 
--threshold_GC 0.8 --symbols_SQ 2 
--seq [16,20] --minimalTest 0 --mode test
--output log_folder/record.txt
\end{verbatim}
In this case, \testRNN\ reads the input LSTM model \param{network} (\param{--model network}) and  generates a test suite of
\num{2}{000} test cases (\param{--TestCaseNum 2000}).
The thresholds for cell and forget gate coverage are 6 and 0.8 (\param{--threshold\_CC 6} \param{--threshold\_GC 0.8}), respectively. 
For sequence coverage, 2 symbols are used for the symbolic representation (\param{--symbols\_SQ 2}) of sequence patterns in cell [19, 24] (\param{--seq [16,20]}). 
The option
\param{--minimalTest 0} stands for not generating the minimal test suite, i.e., all the valid mutated samples will be added to the test suite. A log file \verb!record.txt! is saved to the folder \verb!log_folder!.

\begin{figure}[ht]
    \centering
    \includegraphics[width=0.7\linewidth]{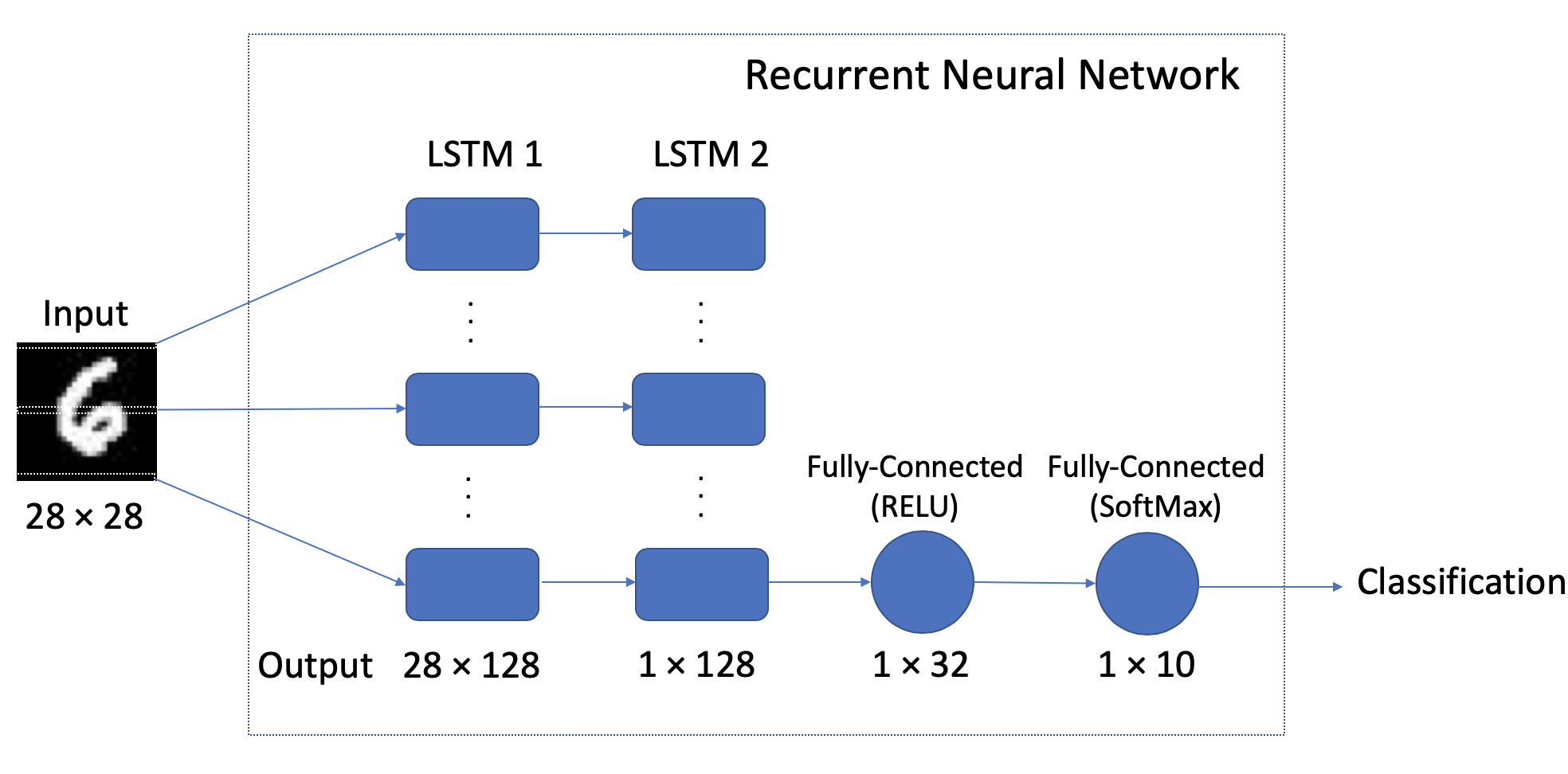}
    \caption{The structure of an
    LSTM network trained on MNIST database.}
    \label{fig:mnist_rnn}
    \vspace{-10pt}
\end{figure}

\section{Experiments}\label{sec:experiments}

We demonstrate the utility of \testRNN\ using a simple LSTM model trained for MNIST handwritten digit recognition. 
Exemplar usages of \testRNN\  on two other LSTM models, including a sentiment analysis model and another molecular prediction model can be found online\footnote{\url{https://github.com/TrustAI/testRNN}}.The structure of the MNIST model is shown in Figure~\ref{fig:mnist_rnn}. Each input image is of size $28$$\times$$28$ and the rows of an image are
taken as a sequential input. The first two layers are LSTM layers, 
followed by two fully-connected layers with ReLU and SoftMax activation functions respectively, to process the extracted feature information and output the classification result. The model achieves $99.2\%$ accuracy in training dataset (\num{50}{000} samples) and $98.7\%$ accuracy in test dataset (\num{10}{000} samples). 

\begin{figure}[ht]
    \centering
    \includegraphics[width=0.9\linewidth]{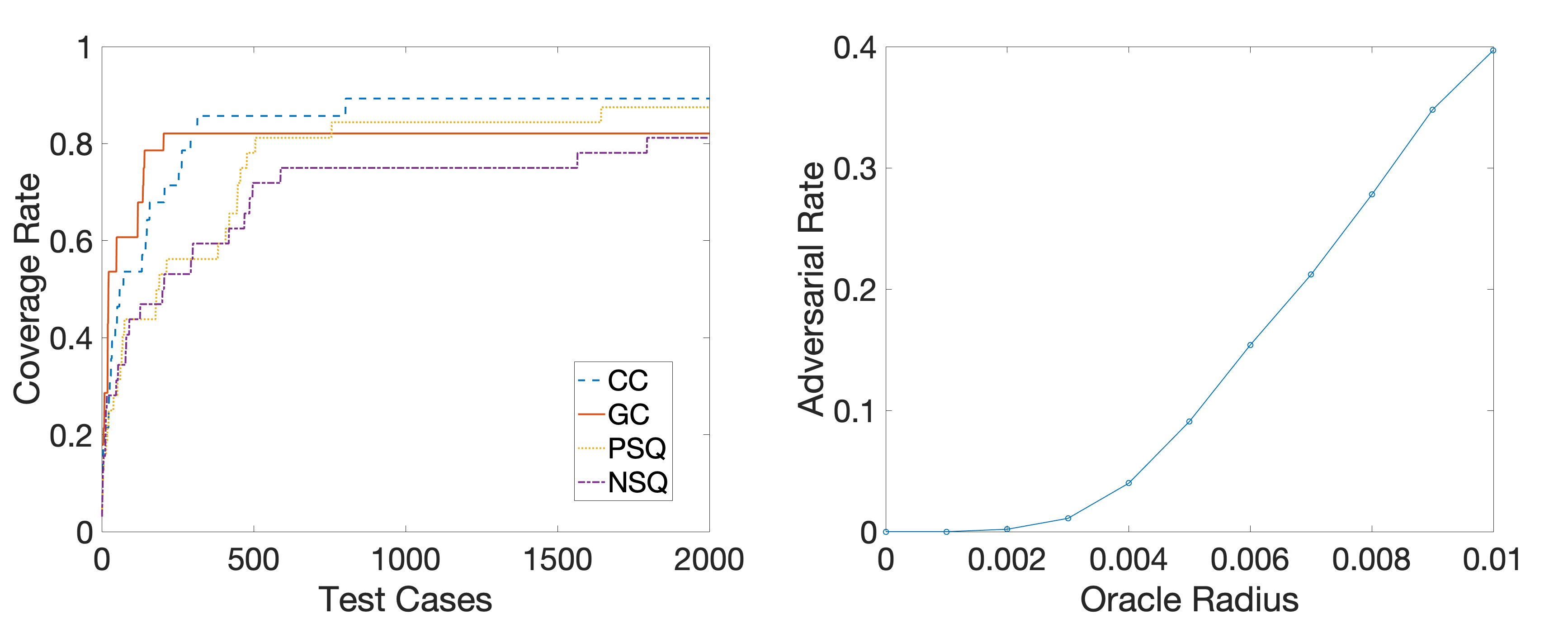}
    \caption{\testRNN\ testing results for the MNIST model.}
    \label{fig:mnist_demo}
    \vspace{-5pt}
\end{figure}

In our experiment, as shown in Fig. \ref{fig:mnist_rnn}, we focus on the LSTM 2 layer. The testing process will terminate when \num{2}{000} test cases are generated. The hyper-parameters set for three test metrics are: $\alpha_h = 6$, $\alpha_f = 0.85$, and $symbols \{a, b\}$. Since each MNIST image has $28$ rows as a sequential input of length $28$, the total number of test-conditions for one-step coverage, including both cell coverage and gate coverage, is $28$. For sequence coverage, for experimental purposes, we let \testRNN\ test on a range of cells, $19$$-$$24$. 
The experiment results in the log file can be visualised as in Figure \ref{fig:mnist_demo} (left) by running the following script:

\begin{verbatim}
python readfile.py --metrcis all 
      --output log_folder/record.txt
\end{verbatim}
It generates the plots of the testing results of all metrics in the report file \verb!log_folder/record.txt!. 

\begin{figure}[ht]
    \centering
    \vspace{-10pt}
    \includegraphics[width=0.8\linewidth]{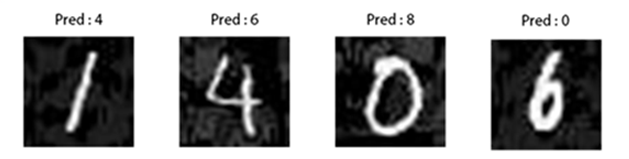}
    \caption{Adversarial Examples For MNIST}
    \label{fig:adv_demo}
    \vspace{-5pt}
\end{figure}

\begin{figure}[ht]
\centering
\includegraphics[width=0.9\linewidth]{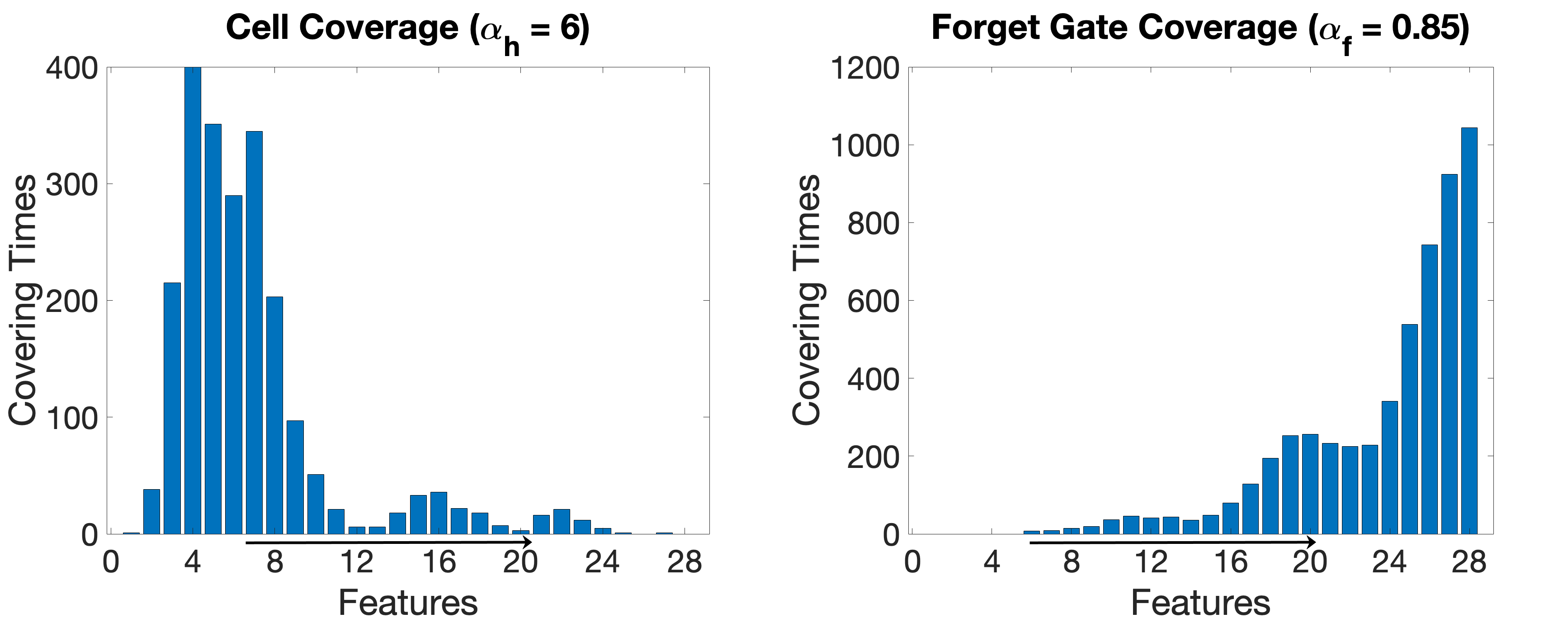}
\caption{2000 test cases are used to demonstrate the coverage times of 28 features in second LSTM layer of the MNIST model.}
\label{fig:mnist_coverage_times}
    \vspace{-10pt}
\end{figure}

Apart from the coverage results for all test metrics, the  plot on the right in Figure \ref{fig:mnist_demo} presents the relationship between the number of adversarial examples in the test suite and the oracle radius. 
We may use the ``area under the curve'' to compare the robustness of networks. Figure~\ref{fig:adv_demo} gives several adversarial examples. 
We can also count the coverage times of all test-conditions for cell and gate coverage. The results displayed in Figure~\ref{fig:mnist_coverage_times} indicate the learning pattern of LSTM for an MNIST model. For example, the cells which are covered more often than others in the left of Figure~\ref{fig:mnist_coverage_times}, such as the 4th cell,  may contribute more towards the final classification.

\section{Conclusions}
\label{sec:concl}

\testRNN\ is a tool for testing and debugging LSTM networks.  We believe that a worthy application of such a white-box testing tool is to
assess the internal structures of an LSTM network, and that this helps 
inform
safety and robustness 
arguments 
for LSTM networks.

\subsubsection*{Acknowledgements}
This document is an overview of UK MOD (part) sponsored research and is released for informational purposes only. The contents of this document should not be interpreted as representing the views of the UK MOD, nor should it be assumed that they reflect any current or future UK MOD policy. The information contained in this document cannot supersede any statutory or contractual requirements or liabilities and is offered without prejudice or commitment.
 
Crown Copyright (2019), Dstl. This material is licensed under the terms of the Open Government Licence except where otherwise stated. To view this licence, visit \url{http://www.nationalarchives.gov.uk/doc/open-government-licence/version/3} or write to the Information Policy Team, The National Archives, Kew, London TW9 4DU, or email: psi@nationalarchives.gsi.gov.uk. 

\bibliography{references}

\begin{thebibliography}{10}

\bibitem{GHTF2004}
G.~S.~G. 3, ``Quality management systems - process validation guidance,'' tech.
  rep., The Global Harmonization Task Force, 2004.

\bibitem{szegedy2014intriguing}
C.~Szegedy, W.~Zaremba, I.~Sutskever, J.~Bruna, D.~Erhan, I.~Goodfellow, and
  R.~Fergus, ``Intriguing properties of neural networks,'' in {\em In ICLR},
  Citeseer, 2014.

\bibitem{YLW2006}
Q.~Yang, J.~J. Li, and D.~Weiss, ``A survey of coverage based testing tools,''
  in {\em Proceedings of the 2006 International Workshop on Automation of
  Software Test}, AST '06, (New York, NY, USA), pp.~99--103, ACM, 2006.

\bibitem{PCYJ2017}
K.~Pei, Y.~Cao, J.~Yang, and S.~Jana, ``{DeepXplore}: Automated whitebox
  testing of deep learning systems,'' in {\em SOSP2017}, pp.~1--18, ACM, 2017.

\bibitem{wicker2018feature}
M.~Wicker, X.~Huang, and M.~Kwiatkowska, ``Feature-guided black-box safety
  testing of deep neural networks,'' in {\em TACAS2018}, pp.~408--426,
  Springer, 2018.

\bibitem{ma2018deepgauge}
L.~Ma, F.~Juefei{-}Xu, J.~Sun, C.~Chen, T.~Su, F.~Zhang, M.~Xue, B.~Li, L.~Li,
  Y.~Liu, J.~Zhao, and Y.~Wang, ``{DeepGauge}: Comprehensive and
  multi-granularity testing criteria for gauging the robustness of deep
  learning systems,'' in {\em ASE2018}, 2018.

\bibitem{sun2018testing}
Y.~Sun, X.~Huang, and D.~Kroening, ``Testing deep neural networks,'' {\em arXiv
  preprint arXiv:1803.04792}, 2018.

\bibitem{sun2018concolic}
Y.~Sun, M.~Wu, W.~Ruan, X.~Huang, M.~Kwiatkowska, and D.~Kroening, ``Concolic
  testing for deep neural networks,'' in {\em ASE}, 2018.

\bibitem{HSHS2019}
W.~Huang, Y.~Sun, J.~Sharp, and X.~Huang, ``Test metrics for recurrent neural
  networks,'' 2019.

\bibitem{MCDC}
J.~J. {Chilenski} and S.~P. {Miller}, ``Applicability of modified
  condition/decision coverage to software testing,'' {\em Software Engineering
  Journal}, vol.~9, pp.~193--200, Sep. 1994.

\bibitem{DXLMZL2019}
X.~Du, X.~Xie, Y.~Li, L.~Ma, J.~Zhao, and Y.~Liu, ``Deepcruiser: Automated
  guided testing for stateful deep learning systems,'' {\em CoRR},
  vol.~abs/1812.05339, 2018.

\bibitem{tang2017memory}
Z.~Tang, Y.~Shi, D.~Wang, Y.~Feng, and S.~Zhang, ``Memory visualization for
  gated recurrent neural networks in speech recognition,'' in {\em ICASSP2017},
  pp.~2736--2740, IEEE, 2017.

\bibitem{goodfellow2014explaining}
I.~J. Goodfellow, J.~Shlens, and C.~Szegedy, ``Explaining and harnessing
  adversarial examples,'' {\em arXiv preprint arXiv:1412.6572}, 2014.

\end{thebibliography}
\bibliographystyle{ieeetr}
\end{document}